\documentclass[11pt,a4paper]{article}
\usepackage[hyperref]{acl2021}
\usepackage{times}
\usepackage{latexsym}
\usepackage{subcaption}
\usepackage[T1]{fontenc}
\usepackage[utf8]{inputenc}
\usepackage{microtype}

\usepackage{url}

\usepackage{graphicx}
\usepackage{verbatim}
\usepackage{booktabs}       
\usepackage{amsfonts}       
\usepackage{nicefrac}       
\usepackage{amsmath,amsfonts,bm}
\usepackage{hyperref}
\usepackage{tikz}
\usetikzlibrary{shapes,arrows,fit,positioning,arrows.meta,backgrounds}
\usepackage{pgfplots}
\usepackage{pgfplotstable}
\pgfplotsset{compat=newest}
\usepackage{catchfile}
\usepackage{ifthen}
\usepgfplotslibrary{fillbetween}

\tikzstyle{myarrow} = [>={Stealth[round]},shorten >=1pt,semithick]
\tikzstyle{block} = [draw, inner sep=0.2cm, node distance=0.6cm, fill=yellow!10, font=\small]
\tikzstyle{m} = [draw, inner sep=0.2cm, node distance=0.6cm, fill=blue!10, font=\small]
\tikzstyle{u} = [draw, inner sep=0.2cm, node distance=0.6cm, fill=green!10, font=\small]
\tikzstyle{xblock} = [draw, inner sep=0.2cm, node distance=0.6cm, fill=red!10, font=\small]
\tikzstyle{gray} = [draw, inner sep=0.2cm, node distance=0.6cm, fill=black!03, font=\small]

\usepackage{float}

\hypersetup{
    colorlinks=false,
    linkcolor=blue,
    filecolor=magenta,
    urlcolor=cyan,
}
\def\m{\mathbf{m}}

\def\acc{\text{acc}}

\def\sys/{\textsc{TexRel}}
\def\tre/{\textsc{tre}}
\def\L3{$L^3$}

\tikzstyle{groupblock} = [draw=none, inner sep=0cm, node distance=0.2cm]
\tikzstyle{boundblock} = [draw=none, inner sep=0cm, use as bounding box]

\aclfinalcopy 

\title{\textsc{TexRel}: a Green Family of Datasets for Emergent Communications on Relations}

\author{{\large \bf Hugh Perkins (hp@asapp.com)} \\
  ASAPP (https://asapp.com) \\
  1 World Trade Center, NY 10007 USA
 }

\newenvironment{customlegend}[1][]{%
    \begingroup
    \csname pgfplots@init@cleared@structures\endcsname
    \pgfplotsset{#1}%
}{%
    \csname pgfplots@createlegend\endcsname
    \endgroup
}%

\def\addlegendimage{\csname pgfplots@addlegendimage\endcsname}

\pgfplotsset{
    discard if not two/.style n args={4}{
        x filter/.code={
            \edef\tempa{\thisrow{#1}}
            \edef\tempb{#2}
            \ifx\tempa\tempb
                \edef\tempa{\thisrow{#3}}
                \edef\tempb{#4}
                \ifx\tempa\tempb
                \else
                    
                \fi
            \else
                
            \fi
        }
    }
}

\begin{document}

\maketitle

\begin{abstract}
We propose a new dataset \sys/ as a playground for the study of emergent communications, in particular for relations. By comparison with other relations datasets, \sys/  provides rapid training and experimentation, whilst being sufficiently large to avoid overfitting in the context of emergent communications. By comparison with using symbolic inputs, \sys/ provides a more realistic alternative whilst remaining efficient and fast to learn. We compare the performance of \sys/ with a related relations dataset Shapeworld. We provide baseline performance results on \sys/ for sender architectures, receiver architectures and end-to-end architectures. We examine the effect of multitask learning in the context of shapes, colors and relations on accuracy, topological similarity and clustering precision. We investigate whether increasing the size of the latent meaning space improves metrics of compositionality. We carry out a case-study on using \sys/ to reproduce the results of an experiment in a recent paper that used symbolic inputs, but using our own non-symbolic inputs, from \sys/, instead.



\end{abstract}

\section{Introduction}

Emergent communications is the study of the linguistic behavior of agents with no pre-training on natural human languages, when placed in a situation where inter-agent communications is needed in order to maximize performance. We can investigate characteristics of the resulting language, such as compositionality; and the extent to which the agents learn to communicate at all. Agents can learn to communicate pictures to each other, e.g. \citet{lazaridou2018_refgames}, or to negotiate with each other, e.g. \citet{cao2018emergent}. In general, the resulting emergent language has limited compositionality. For example, \citet{lazaridou2018_refgames} presented results showing that even when the agents have learned to solve a task with 98\% accuracy, the topographic similarity - a measure of compositionality - might be only around 0.16-0.26. The resulting languages do not tend to clearly show certain key characteristics of human languages, such as the formation of atomic re-usable units of tokens, i.e. words. 

\begin{table}[t]
    \small
    \centering
    \begin{tabular}{p{0.04\textwidth}p{0.06\textwidth}p{0.14\textwidth}p{0.12\textwidth}}
    \toprule
        Num attrs & Num values & Total size & Factorized size  \\
    \midrule
        1 & 3 & $3^1=3$ & $3 * 1 = 3$ \\
        1 & 10 & $10^1=10$ & $10 * 1 = 10$ \\
        2 & 10 & $10^2=100$ & $10 * 2 = 20$ \\
        3 & 10 & $10^3=1{,}000$ & $10 * 3 = 30$ \\
        4 & 10 & $10^4=10{,}000$ & $10 * 4 = 40$ \\
        5 & 10 & $10^5=100{,}000$ & $10 * 5 = 50$ \\
    \bottomrule
    \end{tabular}
    \caption{Comparison of total meaning space size with factorized meaning space size, for various configurations of number of attributes, and values per attribute}
    \label{tab:factorized_atts_values}
\end{table}

\input{figures/texrel_dataset}

We hypothesize that in order to increase the compositionality of the emergent languages, we need to increase the dimensionality of the underlying meaning space, such that the only feasible way for models to be able to store the language is to store it in factorized form. For example, if a language has 10 words for colors, and 10 words for shapes, then a model need memorize only these 20 words in order to describe all possible combinations of colors and shapes. However, if a model uses a unique non-compositional word for each combination of colors and shapes, then the model will need to memorize $10^2=100$ such words, which is a heavier burden. Table \ref{tab:factorized_atts_values} shows a comparison of the total size of the meaning space, and the size required to store a language in compositional, that is `factorized', form. We can see that for 2 attributes, the space required to store a language is comparable for compositional vs non-compositional. As the number of attributes increases to 4 and above, the space required to store a holistic language, that is a non-compositional language, becomes orders of magnitudes higher than for compositional form, given 10 values per attribute.

Thus, a key step to increasing the compositionality of emergent languages is to train agents in an environment of sufficient complexity, that is with many meaning space dimensions. Our work presents a dataset, \sys/, which provides an experimental playground for learning emergent languages in a relatively high dimensional meaning space. In this work, we experiment with meaning spaces with up to 6 dimensions.

We find counter-intuitively that increasing the dimensionality of the meaning space does not increase traditional metrics of compositionality, such as topographic similarity ($\rho$, \citep{brighton_kirby_2006_topographic_mappings}) nor of more recent compositional metrics such as \textsc{tre} \citep{andreas_measuring_compositionality_2019}. Whilst this could show that increasing the meaning space does not increase underlying latent ground truth compositionality, we argue that our counter-intuitive result might instead be because existing compositionality metrics do not correlate perfectly with underlying latent ground-truth compositionality. Thus our results show that there could be an opportunity to develop new compositionality metrics, or to refine existing ones.

One way to create a high-dimensional dataset is to use symbolic inputs, e.g. Study 1 in \citet{lazaridou2018_refgames}. However, symbolic input is essentially a language in compositional form, where each token is a single word describing one attribute. Thus it is unclear whether any emergent compositional language is a reflection of a tendancy of the agents to learn compositional representations, or to simply reflect the compositional representation of the input.

An alternative approach, which we use in this work, is to represent meanings using images. Each image contains one or more objects, each having shape and color. In addition, we can use the relative positioning of two objects to express relations between objects, adding an additional meaning dimension. An existing dataset, Shapeworld, \citet{andreas-etal-2018-learning}, provides such a dataset. However, as we shall see the training set is small, and models capable of learning on the training set quickly overfit, as alluded to in \citet{andreas-etal-2018-learning}. \sys/ provides a much larger training set, 100k training examples, each with 256 images, compared to 9k training images, each with 6 images.

We seek an experimental playground for emergent communications which not only provides high dimensional meaning spaces, using non-symbolic input, but which should ideally be relatively fast to train on. We seek thus to provide images of relatively low resolution, which are friendly to convolutional networks. We note from e.g. \citet{leopardspotsofa} that convolutional networks might pay more attention to the textures of objects than to their outline shape. We thus generate textured objects, all of the same shape, that is a square, rather than shapes with differing outlines, but identical textures. Shapeworld by comparison provides objects with identical solid texture, differing in outline shape. We argue that our approach of using textures allows the use of lower dimensional images, which are easier for a convolutional network to learn. Thus we argue that our family of datasets is `green', that is easy to use in relatively low computational resource environments.

We demonstrate our dataset using a sequence of experiments. We start by comparing the results of training agents to learn a language on Shapeworld versus \sys/ datasets. We examine several potential architectures for the sender agent and for the receiver agent, and measure their performance on \sys/. We investigate the effect of the number of attributes, and the number of values per attribute, on metrics of compositionality. We look at the extent to which multitask learning across different tasks improves performance on the target task. Finally, we provide a case-study of taking an existing work which uses symbolic inputs, and reproducing the work using non-symbolic inputs, using \sys/.

Our contributions are as follows:

\begin{itemize}
    \item create a new dataset, \sys/, which provides a playground for emergent communications
    \begin{itemize}
        \item uses non-symbolic inputs, i.e. images
        \item is fast to train on, having low-dimensional images, where shapes are distinguished by texture, rather than by outline
        \item is much larger than comparable existing emergent communications datasets
        \item provides a high dimensional underlying meaning space, allowing experimentation on how the dimensionality of meaning space affects compositionality
    \end{itemize}
    \item we provide extensive baselines and empirical studies using \sys/:
    \begin{itemize}
        \item compare \sys/ with Shapeworld
        \item compare potential sender agent and receiver agent architectures
        \item examine the effect of meaning space size and dimensionality on compositionality
        \item examine the effect of multi-task learning on learning and compositionality, in the context of \sys/
        \item provide a case-study of using \sys/ in place of symbolic inputs, for fast experimentation, on non-symbolic inputs
    \end{itemize}
\end{itemize}

\section{Our work}

\subsection{Relations learning in the context of emergent communications}

\tikzstyle{myarrow} = [>={Stealth[round]},shorten >=1pt]
\tikzstyle{block} = [draw, inner sep=0.1cm, node distance=0.3cm, fill=yellow!10, font=\small]
\tikzstyle{blueblock} = [draw, inner sep=0.1cm, node distance=0.3cm, fill=blue!10, font=\small]
\tikzstyle{greenblock} = [draw, inner sep=0.1cm, node distance=0.3cm, fill=green!10, font=\small]
\tikzstyle{pinkblock} = [draw, inner sep=0.1cm, node distance=0.3cm, fill=red!10, font=\small]
\tikzstyle{gray} = [draw, inner sep=0.1cm, node distance=0.3cm, fill=black!03, font=\small]

\begin{figure}[]

\begin{subfigure}{0.4\textwidth}
\centering
\begin{tikzpicture}
    \node [block, text width=1.1cm, align=center] (sender) {Sender Agent};
    \node [blueblock, above=of sender] (X_train) {$X_{trn}$};
    \node [blueblock, left=of sender] (Y_train) {$Y_{trn}$};

    \node [greenblock, right=of sender] (utterance) {$\hat{u}$};

    \node [block, text width=1.3cm, align=center, right=of utterance] (receiver) {Receiver Agent};
    \node [blueblock, above=of receiver] (X_test) {$X_{tst}$};
    \node [blueblock, right=of receiver] (Y_test_pred) {$\hat{Y}_{tst}$};
    
    \draw [myarrow, ->] (X_train) -- (sender);
    \draw [myarrow, ->] (Y_train) -- (sender);
    \draw [myarrow, ->] (sender) -- (utterance);
    \draw [myarrow, ->] (utterance) -- (receiver);
    \draw [myarrow, ->] (X_test) -- (receiver);
    \draw [myarrow, ->] (receiver) -- (Y_test_pred);
\end{tikzpicture}
\caption{Emergent communications evaluation}
\end{subfigure}

\begin{subfigure}{0.4\textwidth}
\centering
\begin{tikzpicture}[]
    \node [block] (model) {Model};
    
    \node [blueblock, above left=of model] (X_train) {$X_{trn}$};
    \node [blueblock, , left=of model] (Y_train) {$Y_{trn}$};
    \node [blueblock, above right=of model] (X_test) {$X_{tst}$};
    \node [blueblock, , right=of model] (Y_test_pred) {$\hat{Y}_{tst}$};
    
    \draw [myarrow, ->] (X_train) -- (model);
    \draw [myarrow, ->] (Y_train) -- (model);
    \draw [myarrow, ->] (X_test) -- (model);
    \draw [myarrow, ->] (model) -- (Y_test_pred);
\end{tikzpicture}
\caption{Meta-learning evaluation}
\end{subfigure}

\caption{Emergent communications referential task evaluation, and relationship to meta-learning}
\label{fig:emergent_comms_eval}
\end{figure}
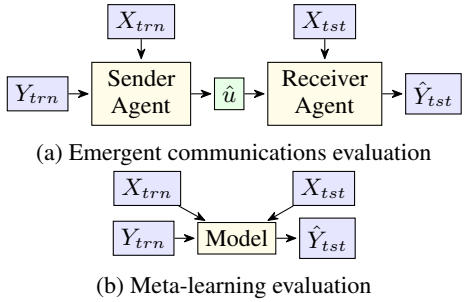

\sys/ targets a referential task, e.g. \citet{lazaridou2018_refgames}, in the context of emergent communications. See Figure \ref{fig:emergent_comms_eval} (a): a sender agent receives labeled sender images for an example and emits a linguistic utterance. A receiver agent receives the linguistic utterance from the sender agent, along with unlabeled receiver images, and is tasked with predicting the correct label (True or False) for each of the receiver images. That is, the Receiver agent needs to predict which of the receiver images are correct examples of the underlying concept. We hope tha the Sender agent will learn to represent the underlying concept from the sender images in the generated linguistic utterance. In the figure, $X_{trn}$ represents the sender images, $Y_{trn}$ represents their labels, and $X_{tst}$ represents the images provided to the receiver. The Receiver agent predicts $\hat{Y}_{tst}$ which will be compared to the ground truth $Y_{tst}$. $\hat{u}$ represents the linguistic utterance generated by the Receiver agent. By comparison with (b), we can see a referential task relates to meta-learning, where the Model has been partitioned into two parts, and a linguistic bottleneck inserted between the two halves.

In practice, in many emergent communications works $X_{trn}$ comprises a single image, and $Y_{trn}$ is always True, i.e. the single image is a positive example of the underlying concept. However, in our work, we wish to target relations, and other concepts that cannot be expressed by a single positive example. Therefore, we have generalized the sender images to be potentially multiple images, potentially both positive and negative, as shown in Figure \ref{fig:emergent_comms_eval} (a). Given the similarity of the referential task for relations to meta-learning, we continue to include multiple receiver examples, in line with standard meta-learning tasks. We choose to draw $X_{tst}$ and $Y_{tst}$ from the same distributions as $X_{trn}$ and $Y_{trn}$. That is, each element of $Y_{tst}$ is a binary True/False value: the Receiver agent needs to label each image in $X_{tst}$ as a true or false example of the underlying concept. This contrasts with many referential tasks where the Receiver agent should choose the one true example of the concept from many distractors, i.e. a multiclass problem.

\subsection{Task types}

\begin{table}[]
    \centering
    \small
    \begin{tabular}{p{0.06\textwidth}p{0.3\textwidth}p{0.04\textwidth}}
    \toprule
    Task type     & Description & Num attrs  \\
    \midrule
     Col$n$ & Includes $n$ objects of colors $\{\mathcal{C}_{c_1}, \dots, \mathcal{C}_{c_n}\}$, where $\{c_1, \dots, c_n\}$ are sampled for each example & $n$ \\
     Tex$n$ & Includes $n$ objects of textures $\{\mathcal{S}_{s_1}, \dots, \mathcal{S}_{s_n}\}$, where $\{s_1, \dots, s_n\}$ are sampled for each example & $n$ \\
     TexCol$n$ & Includes $n$ objects having texture and color $\{(\mathcal{S}_{S_1}, \mathcal{C}_{c_1}), \dots, (\mathcal{S}_{s_n}, \mathcal{C}_{c_n})\}$, where $\{(s_1, c_1), \dots, (s_n, c_n)\}$ are sampled for each example & $2n$ \\
     Rel & Includes an object of texture and color $(\mathcal{S}_{s_1}, \mathcal{C}_{c_1})$ positioned $\mathcal{P}_{p_1}$ relative to an object of texture and color $(\mathcal{S}_{s_2}, \mathcal{C}_{c_2})$, where $(c_1, s_1, p_1, c_2, s_2)$ are sampled for each example & $5$ \\
    \bottomrule
    \end{tabular}
    \caption{Description of each task type in \sys/, where $\mathcal{C}$ is space of available colors, $\mathcal{S}$ is space of available textures, and $\mathcal{P}$ is space of available prepositions. `Num attrs' is the number of attributes, i.e. the dimensionality of the meaning space.}
    \label{tab:task_types}
\end{table}

Table \ref{tab:task_types} shows the task types provided with \sys/. \sys/ can of course easily be extended with additional task types. These task types allow experimentation with varying dimensions of the underlying meaning space. For example, Col3 has a meaning space of 3 dimensions, TexCol3 has a meaning space of 6 dimensions, and Rel has a meaning space of 5 dimensions. We carry out experiments in later sections on the effect of meaning space dimensions on measured compositionality. $\mathcal{C}$ is the space of available colors, where the colors are $\mathcal{C} = \{\mathcal{C}_1, \dots, \mathcal{C}_{N_{colors}}\}$. Similarly $\mathcal{S}$ is the space of textures, and $\mathcal{P}$ is the space of prepositions. Because of symmetry, we include only two prepositions: `above', and `right-of', since the opposite prepositions are indistinguishable visually.

\subsection{Negative examples}

We want to avoid the agents taking short-cuts as much as possible. For example, if the ground truth underlying concept is $(\mathcal{S}_2, \mathcal{C}_4)$, the sender agent could choose to only communicate the texture $\mathcal{S}_2$. If negative examples are sampled uniformly from the space of all possible colors and textures, and the receiver agent eliminates all images not having that texture, then if the number of possible textures is relatively high, then the receiver accuracy will be quite reasonable. To mitigate this possibility somewhat, we keep all negative examples `tight' to the manifold of positive examples. We achieve this by constructing negative examples from positive examples, in which we change just a single attribute. For example, in an example of Col3, each negative example will contain two of the ground truth colors, and only one of the ground truth colors will be missing. Similarly, for an example of TexCol3, each negative example will contain exactly 5 of the ground truth characteristics $\{(\mathcal{S}_{s_1}, \mathcal{C}_{c_1}), \dots, (\mathcal{S}_{s_3}, \mathcal{C}_{c_3}) \}$.

\subsection{Ground truth labels and annotations}

\begin{table*}[t]
    \small
    \centering
    \begin{tabular}{p{0.05\textwidth}p{0.36\textwidth}p{0.5\textwidth}}
    \toprule
        Task & English language annotation & Tree-structured annotation  \\
    \midrule
        Col2 & has-colors color1 color5 & ('has-colors', ('color1', 'color5')) \\
        Tex2 & has-shapes shape2 shape4 & ('has-shapes', ('shape2', 'shape4')) \\
        TexCol2 & has-shapecolors color4 shape1 color5 shape7 & ('has-shapecolors', (('color4', 'shape1'), ('color5', 'shape7'))) \\
        Rel & color0 shape6 above color1 shape6 & ('above', (('color0', 'shape6'), ('color1', 'shape6'))) \\ 
    \bottomrule
    \end{tabular}
    \caption{Examples of English language annotations made available with each \sys/ example.}
    \label{tab:english_language_annotations}
\end{table*}

Each image is associated with a ground truth label, True or False. In addition, for each example, we provide an English language description of the underlying ground truth concept, and a tree-structured representation, that could be used for example with \textsc{tre}. Table \ref{tab:english_language_annotations} shows example annotations.

\subsection{Distractor objects}

We add distractor objects to each image, to increase the size of the state space, and thus aim to discourage the sender from simply sending the entire state of each image. The distractor objects are chosen such that they do not match any of the objects in the underlying ground truth concept for the example; and therefore cannot change a negative example into a positive example, or visa versa.

\subsection{Holdout set}

\begin{table}[]
    \centering
    \small
    \begin{tabular}{ll}
    \toprule
        Task type & Holdout approach \\
        \midrule
        Col$n$ & Set aside several colors \\
        Tex$n$ & Set aside several textures \\
        TexCol$n$ & Set aside several pairs of colors and textures \\
        Rel & Set aside several pairs of colors and textures \\
    \bottomrule
    \end{tabular}
    \caption{Holdout approach by task type.}
    \label{tab:holdout_by_task}
\end{table}

For each task, we carve a holdout set of objects which are not presented at training time. The exact definition of how we decide whether an object was seen at training time or not varies across tasks, see Figures \ref{tab:holdout_by_task}. Following the approach in \citet{andreas-etal-2018-learning}, we name the eval datasets using the training objects `val\_same' and `test\_same', and the eval datasets using the holdout objects `val\_new' and `test\_new'. Given the scarcity of objects in the holdout set for some task types, we draw distractors for `val\_new' and `test\_new` from the union of the training objects and the holdout objects.



\subsection{\sys/ statistics}

For each task, we create a dataset of 100,000 examples. Each example comprises 128 labeled sender images, and 128 labeled receiver images. For each set of 128 labeled images, 64 are positive examples of the underlying concept, and 64 are therefore negative examples. Each evaluation set, i.e. val\_same, val\_new, test\_same and test\_new, has 1024 examples.

\section{Experiments}

\subsection{Experimental setup}

All results are the mean over 5 runs, unless otherwise indicated. Each run used an NVIDIA 2080Ti GPU. Where early stopping was used, we set the patience to 10, and evaluated $\acc_{val\_same}$ set every 300 steps. Batch size was 32, unless otherwise indicated. For the emergent communications linguistic representations, we use a vocab size of 21, and an utterance length of 10.

Code and data will be made available at \footnote{https://github.com/asappresearch/texrel}.

\subsection{Architectures}


\tikzstyle{myarrow} = [>={Stealth[round]},shorten >=1pt]
\tikzstyle{block} = [draw, inner sep=0.1cm, node distance=0.3cm, fill=yellow!10, font=\small]
\tikzstyle{blueblock} = [draw, inner sep=0.1cm, node distance=0.3cm, fill=blue!10, font=\small]
\tikzstyle{greenblock} = [draw, inner sep=0.1cm, node distance=0.3cm, fill=green!10, font=\small]
\tikzstyle{pinkblock} = [draw, inner sep=0.1cm, node distance=0.3cm, fill=red!10, font=\small]
\tikzstyle{gray} = [draw, inner sep=0.1cm, node distance=0.3cm, fill=black!03, font=\small]

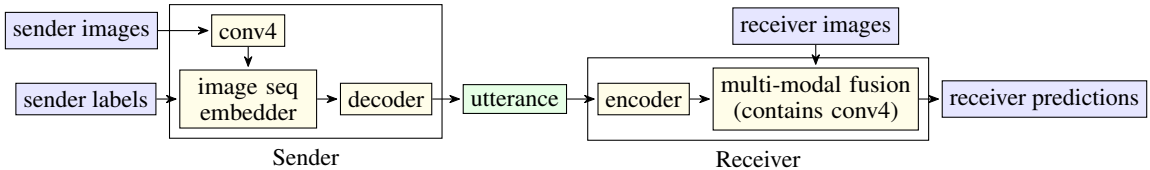
\begin{figure*}[t]
\centering
\begin{tikzpicture}
    \node [block, text width=1.6cm, align=center] (sender) {image seq embedder};
    \node [block, above=of sender] (conv_sender) {conv4};
    \node [blueblock, node distance=0.7cm, left=of conv_sender] (X_train) {sender images};
    \node [blueblock, left=of sender] (Y_train) {sender labels};

    \node [block, right=of sender] (decoder) {decoder};

    \node [greenblock, node distance=0.4cm, right=of decoder] (utterance) {utterance};

    \node [block, node distance=0.4cm, right=of utterance] (encoder) {encoder};

    \node [block, text width=2.5cm, align=center, right=of encoder] (receiver) {multi-modal fusion (contains conv4)};
    \node [blueblock, above=of receiver] (X_test) {receiver images};
    \node [blueblock, right=of receiver] (Y_test_pred) {receiver predictions};
    
    \draw [myarrow, ->] (X_train) -- (conv_sender);
    \draw [myarrow, ->] (conv_sender) -- (sender);
    \draw [myarrow, ->] (Y_train) -- (sender);
    \draw [myarrow, ->] (sender) -- (decoder);
    \draw [myarrow, ->] (decoder) -- (utterance);
    \draw [myarrow, ->] (utterance) -- (encoder);
    \draw [myarrow, ->] (encoder) -- (receiver);
    \draw [myarrow, ->] (X_test) -- (receiver);
    \draw [myarrow, ->] (receiver) -- (Y_test_pred);
\node [draw=black, fit=(sender) (conv_sender) (decoder), label={[font=\small]below:Sender}] {};
\node [draw=black, fit=(receiver) (encoder), label={[font=\small]below:Receiver}] {};
\end{tikzpicture}
\caption{Detailed architecture for sender and receiver models}
\label{fig:sender_receiver_architecture}
\end{figure*}


Figure \ref{fig:sender_receiver_architecture} shows the detailed architecture of the sender and receiver models. For the sender model, in all cases, the model takes as input a sequence of images and labels, fuses these images and labels into an embedding, then uses an RNN linguistic decoder to generate a linguistic utterance. For the receiver model, in all cases, the model takes as input a linguistic utterance and an image, and outputs a prediction of whether the image is consistent with the received linguistic utterance.

\begin{table}[ht]
    \centering
    \small
    \begin{tabular}{p{0.15\textwidth}p{0.3\textwidth}}
\toprule
        Image sequence embedder & Description \\
\midrule
        RNNOverCNN & Encode the images using a CNN, then pass through an RNN \\
        ConvLSTM & LSTM built from convolutions instead of projections \citep{shi2015convolutional} \\
        StackedInputs & Concatenate the image stacks together, along the feature plane dimension, pass through a CNN \\
        MaxPoolingCNN & Pass each image through a CNN, max pool \\
        AveragePoolingCNN & Pass each image through a CNN, average pool \\
        Prototypical & Pass positive images through a CNN, take mean \citep{andreas-etal-2018-learning} \\
\bottomrule
    \end{tabular}
    \caption{Image sequence embedder architectures}
    \label{tab:sender_architectures}
\end{table}

\begin{table}[ht]
    \centering
    \small
    \begin{tabular}{p{0.15\textwidth}p{0.3\textwidth}}
\toprule
        Multi-modal fusion & Description \\
\midrule
        Concat & Concatenate encoded utterance and encoded image, then project, e.g. \citet{misra2017mapping} \\
        Cosine & Dot product of encoded utterance and encoded image, eg \citet{lazaridou2018_refgames} \\
        GatedAtt & Image is encoded using a CNN. Encoded utterance is used as attention over the output planes of the CNN output \citep{Chaplotetal2018} \\
        AllPlaneAtt & Encoded utterance is used to give attention over feature planes of all layers of the CNN, not just the output of the final layer \\
        Configurable Convolution Kernel (`CCK') & Encode utterance, use as weights in convolutional network \citep{chen2015abc} \\
\bottomrule
    \end{tabular}
    \caption{Multi-modal fusion architectures}
    \label{tab:receiver_architectures}
\end{table}

The specific implementations for the image sequence embedder and the multi-modal fusion are described in Tables \ref{tab:sender_architectures} and \ref{tab:receiver_architectures}.  For the sender model, when using the Prototypical sender architecture, only positively labeled images are considered, and other images are ignored. For all other sender architectures, the labels are introduced into the images by adding an additional feature plane to each image, which is either all 1s, or all 0s, depending on the ground truth label for that image. On the receiver side, note that in the general case, the convolutional network cannot be factorized out of the multi-modal fusion. For example, for CCK, the encoded utterance is used as the weights for the convolutional network, and this comprises the entire entanglement between the encoded utterance and the receiver images.

Except where otherwise stated, our experiments use the Prototypical sender architecture and the Cosine receiver architecture. The Cosine receiver architecture is algorithmically identical to the Prototypical receiver architecture described in \citet{andreas-etal-2018-learning} and \citet{mu-etal-2020-shaping}.

The convolutional network architecture is the `conv4' network described in \citet{snell2017prototypical}, which we found worked better than other convolutional architectures we experimented with.


\subsection{Metrics}

We primarily use topographic similarity, which we denote as $\rho$, for measuring the compositionality of utterances, by comparison to a ground-truth description of the underlying concept, e.g. \citet{brighton_kirby_2006_topographic_mappings} and \citet{lazaridou2018_refgames}. In addition, we use holdout accuracy on unseen objects, which we denote as $\acc_{new}$, $\acc_{val\_new}$, or $\acc_{test\_new}$, as an indication of the compositionality of the agents and their communication as a whole. For example, if the agents have seen a $(\mathcal{S}_3, \mathcal{C}_7)$, and a $(\mathcal{S}_2, \mathcal{C}_5)$, then if they do well on a $(\mathcal{S}_3, \mathcal{C}_5))$, then that suggests that their processing of the texture and the color, including their communications about these, are somewhat orthogonal, and thus compositional. However, note that a high $\acc_{new}$ potentially might indicate little about the apparent compositional structure of the emitted linguistic utterance, since the utterance might have been projected arbitrarily, e.g. see \citet{locatello2019challenging}.

We would like to measure also the expressivity and consistency of the generated language. Expressivity is the extent to which the generated language can express all meanings, e.g \citet{kirby2015compression}. Consistency is the extent to which meanings map to the same utterance, e.g. \citet{dagan2020co}. One approach to measuring expressivity is to measure the number of unique utterances in the language, the lexicon size, as per \citet{lazaridou2018_refgames}. However the lexicon size does not take into account the number of unique ground truth hypotheses. We borrow two metrics from cluster analysis: cluster precision and cluster recall. We group each set of identical generated utterances as predicted clusters; and each set of identical ground truth hypotheses as ground truth clusters. Then, for each possible pair of examples, we evaluate whether the examples are in the same predicted cluster (`positive'), or not (`negative'), and similarly for the ground truth clusters, thus classifying each pair as true/false positive or true/false negative. Then we calculate cluster precision as $tp/(tp+fp)$ and cluster recall as $tp/(tp+fn)$. Cluster precision is high when each unique generated utterance maps to a single ground truth hypotheses. Cluster precision is thus a measure of expressivity. Similarly, recall is high when each unique ground truth utterance maps to a single generated utterance. Recall is thus a measure of consistency. Note that using cluster recall to measure consistency is not the only possible approach: \citet{dagan2020co} use Jaccard Similarity instead. In our experiments we find that precision is surprisingly low: language expressivity is systematically low; while recall is systematically high. Thus the models are generating language with good consistency, but with insufficiently diversity to cover the entire meaning space.

A recent metric of compositionality is \textsc{tre}, which measures the extent to which an evaluation model taking as input a ground truth concept can generate the emitted linguistic utterances, under certain compositional constraints. We experiment with \textsc{tre} in a later experiment in this work.


\subsection{Comparison with Shapeworld dataset}

\begin{table*}[htb!]
\small
\centering
\begin{tabular}{ llllllllllllll }
\toprule
 &  & \multicolumn{2}{l}{Sampler} & Time & train & \multicolumn{4}{l}{test\_same} & \multicolumn{4}{l}{test\_new} \\
Code & Dataset & train & eval & (mins) & $\acc$ & $\acc$ & $\rho$ & prec & rec & $\acc$ & $\rho$ & prec & rec \\ 
\midrule
LSL & Shapeworld & soft & soft & 7 & \textbf{0.77} & 0.49 &  &  &  & 0.50 &  &  &  \\ 
LSL & Shapeworld & soft & discr & 15 & 0.62 & 0.50 &  &  &  & 0.50 &  &  &  \\ 
LSL & Shapeworld+aug & soft & soft & 2 & 0.51 & 0.50 &  &  &  & 0.50 &  &  &  \\ 
LSL & Shapeworld+aug & soft & discr & 14 & 0.47 & 0.50 &  &  &  & 0.50 &  &  &  \\ 
ours & Shapeworld & soft & soft & 5 & 0.69 & 0.53 & 0.01 & \textbf{0.06} & 0.7 & 0.52 & 0.01 & 0.07 & 0.5 \\ 
ours & Shapeworld & gumb & discr & 4 & 0.57 & 0.51 & 0.01 & 0.01 & 0.7 & 0.52 & 0.01 & 0.01 & 0.6 \\ 
ours & \textsc{TexRel} & soft & soft & 50 & 0.76 & \textbf{0.75} & \textbf{0.10} & \textbf{0.06} & 0.98 & \textbf{0.68} & \textbf{0.19} & \textbf{0.14} & 0.77 \\ 
ours & \textsc{TexRel} & gumb & discr & 40 & 0.68 & 0.67 & 0.08 & 0.03 & \textbf{0.99} & 0.63 & 0.15 & 0.05 & \textbf{0.81} \\ 
\bottomrule
\end{tabular}
\caption{Comparison between \textsc{{TexRel}} and ShapeWorld datasets for emergent communications scenario. Each result is mean over 5 runs. `soft' means Softmax, `gumb' means Gumbel, `discr` means sample from a Categorical distribution. `+aug' denotes with data augmentation.}
\label{tab:vs_shapeworld}
\end{table*}

\tikzstyle{myarrow} = [>={Stealth[round]},shorten >=1pt]
\tikzstyle{block} = [draw, inner sep=0.1cm, node distance=0.3cm, fill=yellow!10, font=\small]
\tikzstyle{blueblock} = [draw, inner sep=0.1cm, node distance=0.3cm, fill=blue!10, font=\small]
\tikzstyle{greenblock} = [draw, inner sep=0.1cm, node distance=0.3cm, fill=green!10, font=\small]
\tikzstyle{pinkblock} = [draw, inner sep=0.1cm, node distance=0.3cm, fill=red!10, font=\small]
\tikzstyle{gray} = [draw, inner sep=0.1cm, node distance=0.3cm, fill=black!03, font=\small]

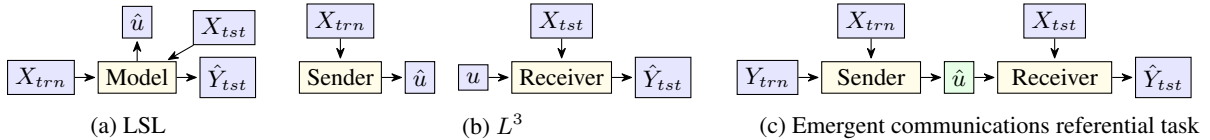
\begin{figure*}
\centering
\begin{subfigure}[t]{0.2\textwidth}
\begin{tikzpicture}
    \node [block] (model) {Model};
    
    \node [blueblock, left=of model] (X_train) {$X_{trn}$};
    \node [blueblock, above right=of model] (X_test) {$X_{tst}$};
    \node [blueblock, above=of model] (utterance) {$\hat{u}$};
    \node [blueblock, right=of model] (Y_test_pred) {$\hat{Y}_{tst}$};

    \draw [myarrow, ->] (X_train) -- (model);
    \draw [myarrow, ->] (X_test) -- (model);
    \draw [myarrow, ->] (model) -- (Y_test_pred);
    \draw [myarrow, ->] (model) -- (utterance);

\end{tikzpicture}
\caption{LSL}
\end{subfigure}
\hfill
\begin{subfigure}[t]{0.32\textwidth}
\begin{tikzpicture}

    \node [block] (sender) {Sender};
    \node [blueblock, above=of sender] (X_train) {$X_{trn}$};

    \node [blueblock, right=of sender] (utteranceleft) {$\hat{u}$};
    \node [blueblock, right=of utteranceleft] (utteranceright) {$u$};

    \node [block, right=of utteranceright] (receiver) {Receiver};
    \node [blueblock, above=of receiver] (X_test) {$X_{tst}$};
    \node [blueblock, right=of receiver] (Y_test_pred) {$\hat{Y}_{tst}$};

    \draw [myarrow, ->] (X_train) -- (sender);
    \draw [myarrow, ->] (sender) -- (utteranceleft);

    \draw [myarrow, ->] (utteranceright) -- (receiver);
    \draw [myarrow, ->] (X_test) -- (receiver);
    \draw [myarrow, ->] (receiver) -- (Y_test_pred);

\end{tikzpicture}
\caption{\L3}
\end{subfigure}
\hfill
\begin{subfigure}[t]{0.4\textwidth}
\begin{tikzpicture}
    \node [block, text width=1.1cm, align=center] (sender) {Sender};
    \node [blueblock, above=of sender] (X_train) {$X_{trn}$};
    \node [blueblock, left=of sender] (Y_train) {$Y_{trn}$};

    \node [greenblock, right=of sender] (utterance) {$\hat{u}$};

    \node [block, text width=1.3cm, align=center, right=of utterance] (receiver) {Receiver};
    \node [blueblock, above=of receiver] (X_test) {$X_{tst}$};
    \node [blueblock, right=of receiver] (Y_test_pred) {$\hat{Y}_{tst}$};
    
    \draw [myarrow, ->] (X_train) -- (sender);
    \draw [myarrow, ->] (Y_train) -- (sender);
    \draw [myarrow, ->] (sender) -- (utterance);
    \draw [myarrow, ->] (utterance) -- (receiver);
    \draw [myarrow, ->] (X_test) -- (receiver);
    \draw [myarrow, ->] (receiver) -- (Y_test_pred);
\end{tikzpicture}
\caption{Emergent communications referential task}
\end{subfigure}

\caption{Comparison of training for LSL, \L3 and our emergent communications referential task}
\label{fig:lsl_l3_emergent_comms_training}
\end{figure*}

We start by comparing \sys/ to the existing Shapeworld relations dataset. We use our own code to learn the \sys/ dataset. For Shapeworld, we evaluate using two codebases: a modified version of the implementation of `Learning with Latent Language' (`\L3',  \citet{andreas-etal-2018-learning}) provided with `Learning with Shaped Language' (`LSL', \citet{mu-etal-2020-shaping}); and our own codebase. Figure \ref{fig:lsl_l3_emergent_comms_training} shows how the training of the LSL and \L3 models compares to our emergent communications referential task. We can see that in LSL and \L3, the linguistic utterance is used purely as a supervisory signal during the training. In LSL and \L3, the utterance is provided as a ground-truth, using English language annotations or similar. However, in emergent communications, the utterance is generated, and is not pre-trained or given any other supervision. We thus modified the LSL code to enable end-to-end learning, without any supervised pre-training. The LSL code-base in addition adds data augmentation, which negatively samples data from other concepts. Thus we added also an option to disable this data augmentation. Other training hyper-parameters remained the same: learning took place over 50 epochs, with early stopping on the average of $\acc_{val\_same}$ and $\acc_{val\_new}$, and a batch size of 100. In our own code-base, we used a Prototypical sender architecture, a Cosine receiver architecture, and early stopping based on $\acc_{val\_same}$. A Cosine receiver architecture is algorithmically identical to a Prototypical receiver architecture, as used in the LSL paper. The evaluation in the original LSL code uses discrete sampling of the intermediate latent utterances. We added in addition an option for evaluation using softmax latent utterance representations.

Figure \ref{tab:vs_shapeworld} shows the results. `LSL' code means the modified \L3 implementation provided with the LSL paper. `ours' is our own code-base. The dataset column shows the dataset used. We use the version of Shapeworld provided with the LSL paper. `Shapeworld+aug' denotes Shapeworld dataset with data augmentation. \sys/ uses the Relations task, with 2 distractors. The modified LSL codebase uses soft sampling at training time for the intermediate latent utterances. In our own codebase we provide also Gumbel sampling.

Looking at the table, we can see that the results on Shapeworld dataset using both the LSL codebase, and our own codebase are very similar, and essentially at random chance. The problem is a binary prediction task, so accuracy 0.5 is at chance. In addition, topologic similarity, $\rho$ is near zero. With data augmentation disabled, $\acc_{train}$ reaches about 80\%, but test accuracy is low, so the model is over-fitting to the dataset. With data augmentation enabled, no overfitting occurs, but the model fails to learn to fit to the training set. Compared to this, using the \sys/ dataset, learning fares better. \sys/ still presents a challenging task, with mean accuracy even with soft intermediate utterances not rising above 0.75. However, \sys/ does succeed in learning the tasks somewhat, and we can see that the measure of compositionality, $\rho$ is slightly higher than 0.

We hypothesize that \sys/ dataset enables stronger learning than Shapeworld because of the relative size of the training set, and because the \sys/ shapes are relatively convolutional-network friendly. \sys/ dataset is much bigger than the Shapeworld dataset: 100,000 training examples, each with 128 sender images and 128 receiver images, compared to 9,000 training examples, each with 5 sender images, and 1 receiver image for Shapeworld. The shapes in \sys/ are arguably more similar to textures than the Shapeworld shapes. The shapeworld shapes each have identical solid textures.

Interestingly, both the measure of compositionality ($\rho$), and the measure of expressivity (prec), were higher on `test\_new' than on `test\_same', even though the accuracy was slightly lower.

\subsection{Search for effective sender and receiver architectures}

\begin{table*}[htb!]
\begin{subtable}{\textwidth}
\small
\caption{Effect of sender architecture on $\acc_{test\_same}$.}
\vspace{-0.2cm}
\centering
    \begin{tabular}{ lllllllllll }
    \toprule
Sender architecture & col1 & col2 & col3 & shp1 & shp2 & shp3 & sc1 & sc2 & sc3 & rels \\ 
\midrule 
RNNOverCNN & 0.7 & 0.58 & 0.53 & 0.64 & 0.56 & 0.53 & 0.49 & 0.33 & 0.28 & 0.33 \\ 
ConvLSTM & 0.54 & 0.47 & 0.52 & 0.56 & 0.55 & 0.51 & 0.51 & 0.35 & 0.29 & 0.34 \\ 
StackedInputs & 0.80 & 0.78 & 0.75 & 0.83 & \textbf{1.00} & \textbf{1.00} & \textbf{0.97} & 0.61 & 0.50 & 0.98 \\ 
MaxPoolingCNN & 0.9 & 0.9 & 0.9 & 0.97 & 0.8 & 0.99 & 0.73 & 0.60 & \textbf{0.53} & 0.86 \\ 
AveragePoolingCNN & \textbf{0.96} & 0.88 & 0.96 & 0.9 & 0.77 & 0.85 & 0.90 & \textbf{0.64} & \textbf{0.53} & 0.88 \\ 
PrototypicalSender & 0.9 & \textbf{1.00} & \textbf{1.00} & \textbf{1.00} & \textbf{1.00} & 0.8 & \textbf{0.97} & \textbf{0.64} & \textbf{0.53} & \textbf{0.99} \\ 
\bottomrule
\end{tabular}
\end{subtable}
\vspace{0.2cm}

\begin{subtable}{\textwidth}
\small
\caption{Effect of sender architecture on $\acc_{test\_new}$.}
\vspace{-0.2cm}
\centering
    \begin{tabular}{ lllllllllll }
    \toprule
Sender architecture & col1 & col2 & col3 & shp1 & shp2 & shp3 & sc1 & sc2 & sc3 & rels \\ 
\midrule 
RNNOverCNN & 0.50 & 0.46 & 0.47 & 0.50 & 0.45 & 0.48 & 0.39 & 0.23 & 0.20 & 0.22 \\ 
ConvLSTM & 0.50 & 0.41 & 0.43 & 0.50 & 0.42 & 0.46 & 0.41 & 0.29 & 0.22 & 0.25 \\ 
StackedInputs & 0.50 & 0.46 & 0.57 & 0.50 & 0.56 & 0.91 & 0.74 & 0.59 & 0.40 & \textbf{0.84} \\ 
MaxPoolingCNN & 0.50 & 0.51 & 0.7 & 0.50 & 0.57 & \textbf{0.98} & 0.59 & 0.53 & 0.48 & 0.50 \\ 
AveragePoolingCNN & 0.50 & 0.57 & 0.91 & 0.50 & 0.55 & 0.83 & 0.58 & 0.57 & 0.52 & 0.58 \\ 
PrototypicalSender & 0.50 & \textbf{0.77} & \textbf{0.98} & 0.50 & \textbf{0.66} & 0.8 & \textbf{0.79} & \textbf{0.64} & \textbf{0.55} & 0.78 \\ 
\bottomrule
\end{tabular}
\end{subtable}
\vspace{0.2cm}

\begin{subtable}{\textwidth}
\small
\caption{Effect of receiver architecture on $\acc_{test\_same}$.}
\vspace{-0.2cm}
\centering
    \begin{tabular}{ lllllllllll }
    \toprule
Receiver architecture & col1 & col2 & col3 & shp1 & shp2 & shp3 & sc1 & sc2 & sc3 & rels \\ 
\midrule 
Concat & 0.61 & 0.50 & 0.51 & 0.60 & 0.53 & 0.51 & 0.50 & 0.50 & 0.50 & 0.50 \\ 
Cosine & \textbf{1.00} & \textbf{1.00} & \textbf{1.00} & 0.96 & 0.99 & \textbf{1.00} & 0.98 & 0.84 & 0.75 & 0.84 \\ 
GatedAtt & 0.99 & \textbf{1.00} & \textbf{1.00} & \textbf{1.00} & \textbf{1.00} & 0.98 & 0.98 & 0.84 & 0.73 & 0.78 \\ 
AllPlaneAtt & \textbf{1.00} & \textbf{1.00} & 0.99 & \textbf{1.00} & \textbf{1.00} & \textbf{1.00} & \textbf{1.00} & \textbf{0.99} & \textbf{0.83} & \textbf{1.00} \\ 
CCK & 0.56 & 0.52 & 0.50 & 0.50 & 0.51 & 0.50 & 0.50 & 0.50 & 0.50 & 0.50 \\ 
\bottomrule
\end{tabular}
\end{subtable}
\vspace{0.2cm}

\begin{subtable}{\textwidth}
\small
\caption{Effect of receiver architecture on $\acc_{test\_new}$.}
\vspace{-0.2cm}
\centering
    \begin{tabular}{ lllllllllll }
    \toprule
Receiver architecture & col1 & col2 & col3 & shp1 & shp2 & shp3 & sc1 & sc2 & sc3 & rels \\ 
\midrule 
Concat & 0.38 & 0.46 & 0.47 & 0.36 & 0.42 & 0.45 & 0.49 & 0.50 & 0.50 & 0.50 \\ 
Cosine & 0.38 & \textbf{0.99} & 0.97 & 0.41 & 0.83 & 0.96 & 0.80 & 0.74 & \textbf{0.71} & 0.74 \\ 
GatedAtt & 0.40 & 0.96 & 0.97 & 0.37 & \textbf{0.87} & 0.94 & 0.84 & 0.74 & 0.67 & 0.72 \\ 
AllPlaneAtt & 0.50 & 0.83 & \textbf{0.98} & \textbf{0.50} & 0.78 & \textbf{0.99} & \textbf{0.95} & \textbf{0.92} & 0.70 & \textbf{0.98} \\ 
CCK & \textbf{0.58} & 0.49 & 0.52 & \textbf{0.50} & 0.52 & 0.50 & 0.50 & 0.50 & 0.50 & 0.51 \\ 
\bottomrule
\end{tabular}
\end{subtable}

\caption{Comparison of effect of sender and receiver architectures on $\acc_{val\_same}$ and $\acc_{val\_new}$ using \sys/ dataset. `col$n$' is Colors task, where $n$ is the number of entities; `shp$n$' is Shapes task, `sc$n$' is shapes-colors task, and `rels' is Relations task. In all cases, two distractor objects are added to each image. Each result is the mean over 5 runs. All runs are for 5 minutes.}
\label{tab:send_recv_tables}
\end{table*}

We compare the send architectures and received architectures described in Tables \ref{tab:sender_architectures} and \ref{tab:receiver_architectures} respectively. We first train the sender and receiver architectures independently, supervised; then we place the sender and receiver in series, and train end-to-end.

Table \ref{tab:send_recv_tables} (a) and (b) shows the results of training the sender architectures independently, supervised. Each result is the mean of 5 runs, and each run is for 5 minutes elapsed. The Prototypical sender and the StackedInputs model both learn most tasks effectively. We retain these two models for the end-to-end analysis. Interestingly, the simplest models, that is simply passing the feature planes through a CNN then use max pooling or average pooling are very effective. Prototypical model is identical to MaxPoolingCNN, except that negative examples are rejected, whereas MaxPoolingCNN uses all available images, and incorporates the label into the feature planes as an additional layer. Interestingly, Prototypical does slightly better than MaxPoolingCNN, even though it rejects half of the available images. This is perhaps partly because the training in this table is always for 5 minutes elapsed, rather than for a fixed number of steps, so the Prototypical model is exposed to more examples. In addition, perhaps it is easier for the model to learn on only positive examples, rather than to have to learn to interpret the additional label feature planes.

For the receiver models, Table \ref{tab:send_recv_tables} (c) and (d), Cosine, GatedAtt and AllPlaneAtt all learn the data effectively. AllPlaneAtt learns the relations data most effectively. We retain all three models for the end-to-end comparison. Interestingly, CCK entirely failed to learn, in the 5 minutes elapsed available to it. The Concat model also performed at chance.

\begin{table*}[htb!]
\small
\centering
\begin{tabular}{ lllllllllllll }
\toprule
 & &  & Time & train & \multicolumn{4}{l}{test\_same} & \multicolumn{4}{l}{test\_new} \\
Sender & Receiver & Steps & (mins) & acc & acc & $\rho$ & prec & rec & acc & $\rho$ & prec & rec \\ 
\midrule 
StackedInputs & AllPlaneAtt & 3k & 11 & 0.53 & 0.52 & 0.02 & 0.02 & 0.96 & 0.53 & 0.04 & 0.04 & 0.63 \\ 
StackedInputs & Cosine & 16k & 36 & \textbf{0.70} & \textbf{0.68} & \textbf{0.09} & \textbf{0.05} & 0.97 & \textbf{0.65} & \textbf{0.16} & \textbf{0.05} & 0.69 \\ 
StackedInputs & FeatPlaneAtt & 14k & 30 & 0.63 & 0.60 & 0.05 & 0.03 & 0.97 & 0.60 & 0.10 & 0.04 & 0.67 \\ 
Prototypical & AllPlaneAtt & 11k & 30 & 0.67 & 0.63 & 0.06 & 0.02 & 0.98 & 0.57 & 0.11 & 0.04 & 0.80 \\ 
Prototypical & Cosine & 17k & 40 & 0.68 & 0.67 & 0.08 & 0.03 & \textbf{0.99} & 0.63 & 0.15 & \textbf{0.05} & \textbf{0.81} \\ 
Prototypical & FeatPlaneAtt & 17k & 40 & 0.66 & 0.65 & 0.07 & 0.02 & 0.98 & 0.60 & 0.13 & \textbf{0.05} & 0.77 \\  
\bottomrule
\end{tabular}
\caption{Comparison on end-to-end architectures. Each result is mean over 5 runs. Utterances are sampled from Gumbel distributions. The underlying task is Relations. Early stopping on $\acc_{val\_same}$}
\label{tab:e2e_comp}
\end{table*}

Figure \ref{tab:e2e_comp} shows a comparison of all pairs of sender and receiver models retained from the independent supervised sender and receiver model training earlier. This table trains until convergence, using early stopping on $\acc_{val\_same}$. Interestingly, when trained end-to-end, until convergence, AllPlaneAtt performs relatively poorly, even though it fitted best when trained supervised, and generalized well to test\_new. The Cosine receiver performs consistently well, with both the StackedInputs sender, and the Prototypical Sender. The strongest pair of architectures was StackedInputs sender with Cosine receiver. However, the Prototypical sender with the Cosine receiver was also strong.

\subsection{Effect of multi-task training}

\begin{table*}[htb!]
\begin{subtable}{\textwidth}
\small

\caption{$\acc_{test\_same}$}
\vspace{-0.2cm}
\centering
\begin{tabular}{ lllllll }
\toprule
Task & No multitask & Col1,Col2 & Tex1,Tex2 & TexCol1,TexCol2 & all & all, +0 dists \\ 
\midrule 
Col3 & 0.69+/-0.01 & \textbf{0.77+/-0.04} & 0.63+/-0.04 & 0.63+/-0.01 & 0.56+/-0.01 & 0.58+/-0.01 \\ 
Tex3 & 0.62+/-0.02 & 0.67+/-0.02 & \textbf{0.70+/-0.04} & 0.62+/-0.01 & 0.56+/-0.00 & 0.58+/-0.00 \\ 
TexCol3 & 0.59+/-0.02 & 0.67+/-0.02 & \textbf{0.70+/-0.02} & 0.63+/-0.01 & 0.56+/-0.01 & 0.58+/-0.00 \\ 
Rel & 0.59+/-0.01 & \textbf{0.68+/-0.01} & 0.62+/-0.04 & 0.63+/-0.01 & 0.56+/-0.01 & 0.58+/-0.01 \\ 
\bottomrule
\end{tabular}
\end{subtable}
\vspace{0.2cm}

\begin{subtable}{\textwidth}
\small
\caption{$\acc_{test\_new}$}
\vspace{-0.2cm}
\centering
\begin{tabular}{ lllllll }
\toprule
Task & No multitask & Col1,Col2 & Tex1,Tex2 & TexCol1,TexCol2 & all & all, +0 dists \\ 
\midrule 
Col3 & \textbf{0.66+/-0.02} & 0.63+/-0.03 & 0.54+/-0.05 & 0.62+/-0.01 & 0.55+/-0.01 & 0.56+/-0.01 \\ 
Tex3 & 0.61+/-0.02 & 0.53+/-0.01 & 0.58+/-0.02 & \textbf{0.61+/-0.01} & 0.55+/-0.00 & 0.57+/-0.00 \\ 
TexCol3 & 0.56+/-0.01 & 0.59+/-0.05 & 0.61+/-0.04 & \textbf{0.62+/-0.01} & 0.55+/-0.01 & 0.56+/-0.01 \\ 
Rel & 0.58+/-0.01 & 0.56+/-0.01 & 0.58+/-0.04 & \textbf{0.61+/-0.01} & 0.55+/-0.00 & 0.57+/-0.01 \\ 
\bottomrule
\end{tabular}
\end{subtable}
\vspace{0.2cm}

\begin{subtable}{\textwidth}
\small
\caption{$\rho_{test\_same}$}
\vspace{-0.2cm}
\centering
\begin{tabular}{ lllllll }
\toprule
Task & No multitask & Col1,Col2 & Tex1,Tex2 & TexCol1,TexCol2 & all & all, +0 dists \\ 
\midrule 
Col3 & 0.30+/-0.03 & 0.24+/-0.03 & \textbf{0.3+/-0.2} & 0.12+/-0.02 & 0.25+/-0.05 & 0.12+/-0.04 \\ 
Tex3 & 0.18+/-0.05 & \textbf{0.2+/-0.1} & 0.20+/-0.02 & 0.18+/-0.02 & 0.21+/-0.06 & 0.14+/-0.02 \\ 
TexCol3 & 0.07+/-0.01 & \textbf{0.21+/-0.06} & 0.15+/-0.02 & 0.10+/-0.03 & 0.21+/-0.06 & 0.16+/-0.03 \\ 
Rel & 0.07+/-0.01 & 0.17+/-0.04 & 0.18+/-0.07 & 0.06+/-0.01 & \textbf{0.19+/-0.04} & 0.12+/-0.02 \\ 
\bottomrule
\end{tabular}
\end{subtable}
\vspace{0.2cm}

\begin{subtable}{\textwidth}
\small
\caption{$\rho_{test\_new}$}
\vspace{-0.2cm}
\centering
\begin{tabular}{ lllllll }
\toprule
Task & No multitask & Col1,Col2 & Tex1,Tex2 & TexCol1,TexCol2 & all & all, +0 dists \\ 
\midrule 
Col3 & 0.33+/-0.04 & 0.35+/-0.04 & \textbf{0.5+/-0.2} & 0.14+/-0.02 & 0.30+/-0.05 & 0.11+/-0.08 \\ 
Tex3 & 0.15+/-0.03 & \textbf{0.4+/-0.1} & 0.37+/-0.04 & 0.20+/-0.01 & 0.27+/-0.06 & 0.09+/-0.07 \\ 
TexCol3 & 0.11+/-0.05 & \textbf{0.4+/-0.1} & 0.35+/-0.08 & 0.06+/-0.09 & 0.27+/-0.05 & 0.17+/-0.07 \\ 
Rel & 0.11+/-0.02 & 0.36+/-0.05 & \textbf{0.4+/-0.1} & 0.11+/-0.05 & 0.28+/-0.05 & 0.07+/-0.05 \\ 
\bottomrule
\end{tabular}
\end{subtable}

\caption{Effect of multitask training on accuracy and topographic similarity $\rho$. Rows are target tasks; and columns are multi-task training tasks. All are using 0 distractors, except `+0 dists' adds in additional tasks with 0 distractors. All runs use early stopping on $\acc_{val\_same}$}
\label{tab:multitask}
\end{table*}

We wanted to investigate the extent to which multi-task training on simpler related tasks improves performance on a specific target task. Table \ref{tab:multitask} shows the results. The subtables are for $\acc_{test\_same}$, $\acc_{test\_new}$, $\rho_{test\_same}$ and $\rho_{test\_new}$ respectively. Within each table, the rows represent specific target tasks, and the columns are tasks that we use for multi-task training. We train to convergence, using early stopping on $\acc_{val\_same}$ in all cases. `Col1,Col2' denotes multi-task training with the Col1 and Col2 task, and similarly for `Tex1,Tex2', and `TexCol1,TexCol2'. `all` denotes multi-task training on all of `Col1, Col2, Col3, Tex1, Tex2, Tex3, TexCol1, Texcol2, TexCol3, Rel'. For all multi-task training tasks described so far, 2 distractor objects are always added to each image. `all, +0dists' denotes that in addition images with 0 distractor objects are also added.

Looking at $\acc_{test\_new}$ table, we can see that for Col3, none of the multi-task training approaches improves test\_new accuracy. For Tex3, multi-task training on Tex1,Tex2 made the acuracy worse. However, surprisingly, multi-task training on TexCol tasks, whilst not improving the accuracy, did not hurt the accuracy. For TexCol3, and Rel, multi-task training on TexCol1,Texcol2 did improve the test\_new accuracy by several percentage points. The `all` and `all,+0 dists' never improved test\_new accuracy, and almost always harmed it.

Looking at $\acc_{test\_same}$, multi-task training with Col1,Col2, or with Tex1,Tex2 does improve the accuracy. However, as we saw, this accuracy boost did not generalize toe $\acc_{test\_new}$.

Looking at $\rho_{test\_new}$, here the multi-task training does appear to improve the measured compositionality. Multi-task training with Col1,Col2, or with Tex1,Tex2 always improved measured topographic similarity compositionality, across all tasks. Perhaps the multi-task training forces the model to be able to represent more underlying concepts, and thus needs to be able to represent the concepts in a more compositional way, in order to be able to memorize them effectively? This would not be unaligned with the accuracies not improving, since the pressure to learn more underlying concepts might make the task harder to learn, even though the resulting generated linguistic utterances show much stronger evidence of compositionality.

Note that whilst multi task training on `all' did also boost $\rho_{test\_new}$ somewhat, the effect was not as great as using simply Col1,Col2 or Tex1,Tex2. We hypothesize that combining so many tasks together makes the task so challenging that the model fails to learn effectively. And indeed we can see in the $\acc_{test\_new}$ tables that multi-task training on `all' did leed to relatively poor $\acc_{test\_same}$ and $\acc_{test\_new}$.

It looks overall like multi-task training can potentially lead to more compositional utterances. We leave to future work further investigation into the relationship between multi-task training, effective learning of the task, and highly compositional generated language.

\subsection{Effect of size of meaning space $\mathcal{M}$ on metrics of compositionality}

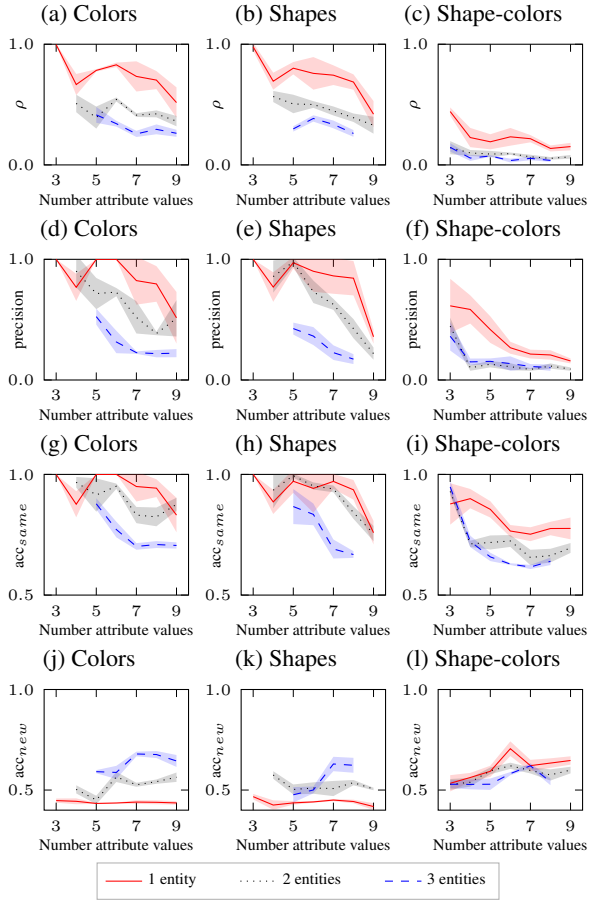
\begin{figure}[t]
\centering
\foreach \valuetype/\valuedesc/\valuedesclong\miny/\ytickdist in {rho/$\rho$/$\rho$/0/1.0, prec/prec/precision/0/1.0, same_acc/$\acc_{sam}$/$\acc_{same}$/0.5/0.5, new_acc/$\acc_{new}$/$\acc_{new}$/0.4/0.5} {
\foreach \entitytype/\entitytypename/\majortickson in {colors/Colors/true,shapes/Shapes/false,things/Shape-colors/false} {
\begin{subfigure}{0.15\textwidth}
\caption{\small{\entitytypename}}
\vspace{-0.2cm}
\begin{tikzpicture}
    \begin{axis}[
        tiny,
        scale only axis,
        width=1.9cm,
        height=1.6cm,
        ylabel=\valuedesclong,
        xlabel=Number attribute values,
        ylabel shift=-0.5cm,
        xlabel shift=-0.15cm,
        yticklabel style={
            /pgf/number format/fixed,
            /pgf/number format/precision=1,
            /pgf/number format/fixed zerofill
        },
        scaled y ticks=false,
        xtick = {3,5,7,9},
        ytick distance=\ytickdist,
        mark size=1pt,
        ymin=\miny,
        ymax=1,
    ]
    \foreach \numentities/\color/\linetype in {1/red/solid, 2/black/dotted, 3/blue/dashed} {
        \edef\tmp{\noexpand\addplot [
            name path=lower,
            draw=none,
        ] table [
            x=num_values,
            y expr=\noexpand\thisrow{\valuetype} - \noexpand\thisrow{\valuetype_ci95},
            discard if not two={num_entities}{\numentities}{entity}{\entitytype},
            col sep=comma,
        ] {data/traa07_11tall.csv}};
        \tmp;
        \edef\tmp{\noexpand\addplot [
            name path=upper,
            draw=none,
        ] table [
            x=num_values,
            y expr=\noexpand\thisrow{\valuetype} + \noexpand\thisrow{\valuetype_ci95},
            discard if not two={num_entities}{\numentities}{entity}{\entitytype},
            col sep=comma,
        ] {data/traa07_11tall.csv}};
        \tmp;
        \edef\tmp{\noexpand\addplot[\color!80, fill opacity=0.2] fill between[of=lower and upper];}
        \tmp;
        \edef\tmp{\noexpand\addplot [
            \color,
            \linetype,
        ] table [
            x=num_values,
            y expr=\noexpand\thisrow{\valuetype},
            discard if not two={num_entities}{\numentities}{entity}{\entitytype},
            col sep=comma,
        ] {data/traa07_11tall.csv}};
        \tmp;
    }
    \end{axis}

\end{tikzpicture}
\end{subfigure}
}
}
\begin{subfigure}{0.5\textwidth}
\centering
\begin{tikzpicture}
    \begin{customlegend}[
        tiny,
        legend entries={1 entity,2 entities,3 entities},
        legend style={
            draw=black!30,
            /tikz/every even column/.append style={column sep=0.5cm},
        },
        legend columns=-1,
    ]
    \addlegendimage{red}
    \addlegendimage{black,dotted}
    \addlegendimage{blue,dashed}
    \end{customlegend}
\end{tikzpicture}
\end{subfigure}

\caption{Experiments on varying size of meaning space $\mathcal{M}$, using \sys/; shaded areas are 95\% confidence intervals, over 5 seeds. Each result is for 5k training steps.}
\label{fig:vary_meaning_space}
\end{figure}

\begin{figure}[htb!]
\centering
\begin{subfigure}{0.22\textwidth}
\caption{}
\begin{tikzpicture}
    \begin{axis}[
        clip=false,
        tiny,
        width=4.2cm,
        height=4.0cm,
        ylabel=generalization error,
        xlabel=\tre/,
        ylabel shift=-0.2cm,
        yticklabel style={
            /pgf/number format/fixed,
            /pgf/number format/precision=2,
            /pgf/number format/fixed zerofill
        },
        scaled y ticks=false,
        xtick distance=0.5,
        ytick distance=0.05,
        mark size=0.5pt,
    ]
    \addplot [only marks, blue] table [x=tre,y=gen_err, col sep=comma] {data/trmc015.csv};
    \addplot [blue] table[
        x=tre,
        y={create col/linear regression = {y = gen_err}}, col sep=comma
    ] {data/trmc015.csv};
    \end{axis}
\end{tikzpicture}
\end{subfigure}
\begin{subfigure}{0.22\textwidth}
\caption{}
\begin{tikzpicture}
    \begin{axis}[
        clip=false,
        tiny,
        width=4.2cm,
        height=4.0cm,
        xlabel=\tre/,
        ylabel=$\acc_{val\_new}$,
        ylabel shift=-0.2cm,
        yticklabel style={
            /pgf/number format/fixed,
            /pgf/number format/precision=2,
            /pgf/number format/fixed zerofill
        },
        scaled y ticks=false,
        xtick distance=0.5,
        ytick distance=0.05,
        mark size=0.5pt,
    ]
    \addplot [only marks, blue] table [x=tre,y=new_acc, col sep=comma] {data/trmc015.csv};
    \addplot [blue] table[
        x=tre,
        y={create col/linear regression = {y = new_acc}}, col sep=comma
    ] {data/trmc015.csv};
    \end{axis}
\end{tikzpicture}
\end{subfigure}

\begin{subfigure}{0.22\textwidth}
\caption{}
\begin{tikzpicture}
    \begin{axis}[
        clip=false,
        tiny,
        width=4.2cm,
        height=4.0cm,
        xlabel=precision,
        ylabel=\tre/,
        ylabel shift=-0.2cm,
        yticklabel style={
            /pgf/number format/fixed,
            /pgf/number format/precision=1,
            /pgf/number format/fixed zerofill
        },
        scaled y ticks=false,
        xticklabel style={
            /pgf/number format/fixed,
            /pgf/number format/precision=2,
            /pgf/number format/fixed zerofill
        },
        scaled x ticks=false,
        xtick distance=0.05,
        ytick distance=0.5,
        mark size=0.5pt,
    ]
    \addplot [only marks, blue] table [x=prec,y=tre, col sep=comma] {data/trmc015.csv};
    \addplot [blue] table[
        x=prec,
        y={create col/linear regression = {y = tre}}, col sep=comma
    ] {data/trmc015.csv};
    \end{axis}
\end{tikzpicture}
\end{subfigure}
\begin{subfigure}{0.22\textwidth}
\caption{}
\begin{tikzpicture}
    \begin{axis}[
        clip=false,
        tiny,
        width=4.2cm,
        height=4.0cm,
        xlabel=$\acc_{val\_same}$,
        ylabel=generalization error,
        ylabel shift=-0.2cm,
        yticklabel style={
            /pgf/number format/fixed,
            /pgf/number format/precision=2,
            /pgf/number format/fixed zerofill
        },
        scaled y ticks=false,
        xticklabel style={
            /pgf/number format/fixed,
            /pgf/number format/precision=2,
            /pgf/number format/fixed zerofill
        },
        scaled x ticks=false,
        xtick distance=0.05,
        ytick distance=0.05,
        mark size=0.5pt,
    ]
    \addplot [only marks, blue] table [x=same_acc,y=gen_err, col sep=comma] {data/trmc015.csv};
    \addplot [blue] table[
        x=same_acc,
        y={create col/linear regression = {y = gen_err}}, col sep=comma
    ] {data/trmc015.csv};
    \end{axis}
\end{tikzpicture}
\end{subfigure}

\begin{subfigure}{0.22\textwidth}
\caption{}
\begin{tikzpicture}
    \begin{axis}[
        clip=false,
        tiny,
        width=4.2cm,
        height=4.0cm,
        xlabel=$\acc_{val\_same}$,
        ylabel=$\acc_{val\_new}$,
        ylabel shift=-0.2cm,
        yticklabel style={
            /pgf/number format/fixed,
            /pgf/number format/precision=2,
            /pgf/number format/fixed zerofill
        },
        scaled y ticks=false,
        xticklabel style={
            /pgf/number format/fixed,
            /pgf/number format/precision=2,
            /pgf/number format/fixed zerofill
        },
        scaled x ticks=false,
        xtick distance=0.05,
        ytick distance=0.05,
        mark size=0.5pt,
    ]
    \addplot [only marks, blue] table [x=same_acc,y=new_acc, col sep=comma] {data/trmc015.csv};
    \addplot [blue] table[
        x=same_acc,
        y={create col/linear regression = {y = new_acc}}, col sep=comma
    ] {data/trmc015.csv};
    \end{axis}
\end{tikzpicture}
\end{subfigure}
\begin{subfigure}{0.22\textwidth}
\caption{}
\begin{tikzpicture}
    \begin{axis}[
        clip=false,
        tiny,
        width=4.2cm,
        height=4.0cm,
        xlabel=precision,
        ylabel=$\acc_{val\_same}$,
        ylabel shift=-0.2cm,
        yticklabel style={
            /pgf/number format/fixed,
            /pgf/number format/precision=2,
            /pgf/number format/fixed zerofill
        },
        scaled y ticks=false,
        xticklabel style={
            /pgf/number format/fixed,
            /pgf/number format/precision=2,
            /pgf/number format/fixed zerofill
        },
        scaled x ticks=false,
        xtick distance=0.05,
        ytick distance=0.05,
        mark size=0.5pt,
    ]
    \addplot [only marks, blue] table [x=prec,y=same_acc, col sep=comma] {data/trmc015.csv};
    \addplot [blue] table[
        x=prec,
        y={create col/linear regression = {y = same_acc}}, col sep=comma
    ] {data/trmc015.csv};
    \end{axis}
\end{tikzpicture}
\end{subfigure}

\begin{subfigure}{0.22\textwidth}
\caption{}
\begin{tikzpicture}
    \begin{axis}[
        clip=false,
        tiny,
        width=4.2cm,
        height=4.0cm,
        xlabel=\textsc{ptre},
        ylabel=generalization error,
        ylabel shift=-0.2cm,
        yticklabel style={
            /pgf/number format/fixed,
            /pgf/number format/precision=2,
            /pgf/number format/fixed zerofill
        },
        scaled y ticks=false,
        xticklabel style={
            /pgf/number format/fixed,
            /pgf/number format/precision=0,
            /pgf/number format/fixed zerofill
        },
        scaled x ticks=false,
        xtick distance=100,
        ytick distance=0.05,
        mark size=0.5pt,
    ]
    \addplot [only marks, blue] table [x=ptre,y=gen_err, col sep=comma] {data/trmc015.csv};
    \addplot [blue] table[
        x=ptre,
        y={create col/linear regression = {y = gen_err}}, col sep=comma
    ] {data/trmc015.csv};
    \end{axis}
\end{tikzpicture}
\end{subfigure}
\begin{subfigure}{0.22\textwidth}
\caption{}
\begin{tikzpicture}
    \begin{axis}[
        clip=false,
        tiny,
        width=4.2cm,
        height=4.0cm,
        xlabel=$\rho$,
        ylabel=generalization error,
        ylabel shift=-0.2cm,
        yticklabel style={
            /pgf/number format/fixed,
            /pgf/number format/precision=2,
            /pgf/number format/fixed zerofill
        },
        scaled y ticks=false,
        xticklabel style={
            /pgf/number format/fixed,
            /pgf/number format/precision=2,
            /pgf/number format/fixed zerofill
        },
        scaled x ticks=false,
        xtick distance=0.05,
        ytick distance=0.05,
        mark size=0.5pt,
    ]
    \addplot [only marks, blue] table [x=rho,y=gen_err, col sep=comma] {data/trmc015.csv};
    \addplot [blue] table[
        x=rho,
        y={create col/linear regression = {y = gen_err}}, col sep=comma
    ] {data/trmc015.csv};
    \end{axis}
\end{tikzpicture}
\end{subfigure}

\caption{Experiments on \tre/ based on \citet{andreas_measuring_compositionality_2019} section 7, using \sys/. Each run is for 5k training steps. Each point represents the result of a single run.}
\label{fig:tre_section7}
\end{figure}
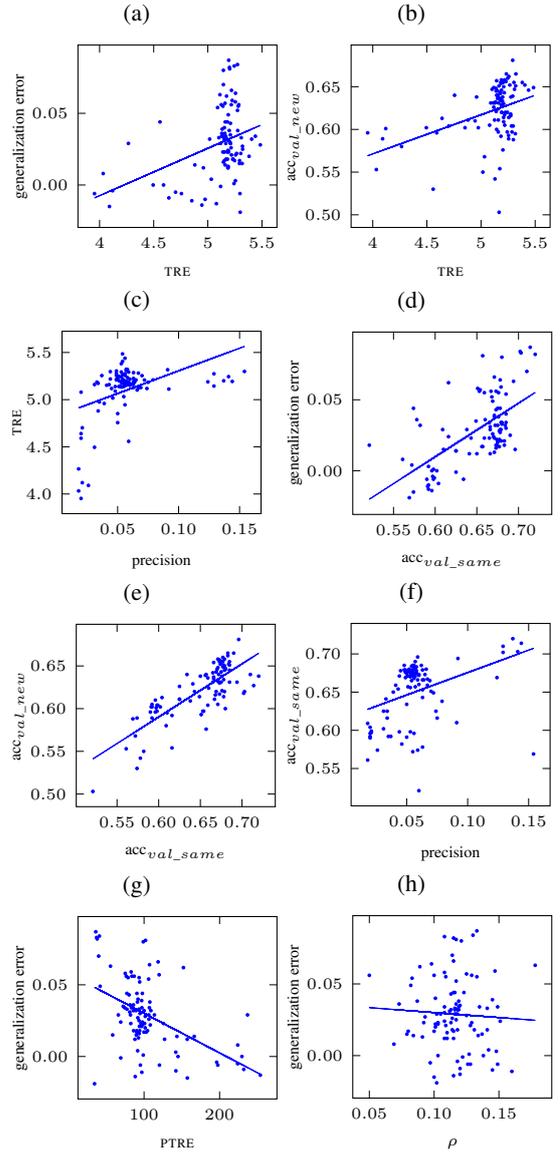

We hypothesized that increasing the number of attributes would increase compositionality, and thus increase $\rho$. To test this, we used datasets for color, shapes, and shape-colors, and varied both the number of entities in the hypothesis, and the number of possible attribute values. Figure \ref{fig:vary_meaning_space} shows the results.

Interestingly, we can see that $\rho$ does not increase as we increase the number of entities or the number of attribute values  (subfigures (a)-(c)). $\rho$ actually decreases, for all conditions. Further, for the case of a single entity, and 3 possible attribute values, both colors and shapes show a $\rho$ of 1.0, meaning perfectly compositional. We note that for the case of a single color or shape entity, given perfect training accuracy, $\rho$ will always be 1.0, even if utterances are drawn randomly for each possible shape or color. Therefore, the measurement of $\rho$ performs perhaps unintuitively at boundary conditions, and there might be an opportunity to find a variation, or alternative measure, of compositionality, that outputs zero for a single attribute value. As the number of attribute values increases, $\acc_{val\_same}$ stays near perfect, at least up to 6-7 values, and yet we can see that $\rho$ falls. It is counter-intuitive that increasing the number of values would decrease underlying compositionality, and indeed $\acc_{new\_acc}$ is consistently at chance, as we would expect. We hypothesize that increasing the number of values decreases $\rho$ for a single color entity and  for a single shape entity because the distances between utterances for different colors is not constant across pairs of differing ground-truth hypotheses, whilst the distance between the pairs of ground-pair hypotheses is always 1, and therefore the Spearman's rank correlation between the two sets of distances decreases from 1. Arguably, for a single entity, changing the number of attribute values does not change the underlying compositionality, since each utterance maps to a single attribute value, and thus $\rho$ changing is arguably not aligned with the underlying ground-truth compositionality, and there could be an opportunity to introduce a modified, or new, metric of compositionality that does not change in this way.

Looking at $\acc_{val\_new}$, we can see that increasing the number of values does improve $\acc_{val\_new}$ sub-figures (j)-(l), in line with our intuition that increasing the number of values will increase compositionality. For Shapes and Colors, the increase becomes stronger as the number of entities in the underlying ground truth hypothesis increases. This is also in line with our hypothesis that increasing the number of attributes will increase compositionality, and despite $\acc_{val\_same}$ actually decreasing with increasing complexity of the task, sub-figures (g)-(i). Thus, there is a contradiction between the compositionality as measured by $\rho$, which decreases with task complexity, and that as measured by $\acc_{val\_new}$, which increases with task complexity. We believe that this shows additional evidence for further work on creating variations on $\rho$, or creating alternative metrics of compositionality.

\subsection{Reproduction of \tre/ Section 7 Experiments using Non-Symbolic Input}

As an example of using \sys/ to run experiments using symbolic data to use non-symbolic data, we reproduce  experiments from section 7 of  \citet{andreas_measuring_compositionality_2019}. We target the experiments that investigate the relationship between compositionality and generalization. Figure \ref{fig:tre_section7} (a) shows the relationship between \tre/ and generalization error, for 100 runs using a Relations dataset from \sys/, drawn from 9 possible shapes and 9 possible colors. We can see that in line with Andreas's results, that generalization error does decrease with lower \tre/, but that in addition $\acc_{val\_new}$ also decreases with decreasing \tre/, Figure \ref{fig:tre_section7} (b). These experiments ran in 17 hours, on a single 2080Ti GPU. Therefore the compact size of the images in \sys/ allows for rapid, low-cost experimentation on non-symbolic image inputs.

As alluded to in \citet{andreas_measuring_compositionality_2019}, we note that low \tre/ correlates with low precision, Figure \ref{fig:tre_section7} (c). The general formula for \tre/ measures the reconstruction error for the utterances, given a ground-truth input. If all utterances are identical, then the \tre/ model can learn to ignore the ground-truth input, and always predict the same utterance, giving zero reconstruction error. We would argue that low values of \tre/ for low precision languages, that is languages with low expressivity, might not indicate high compositionality. We can see that the strongest predictor of generalization error is $\acc_{val\_same}$, Figure \ref{fig:tre_section7} (d): higher $\acc_{val\_same}$ accuracy leads to higher generalization error. We can see that higher $\acc_{val\_same}$ accuracy tends to correlate with higher $\acc_{val\_new}$, Figure \ref{fig:tre_section7} (e), but that the correlation coefficient is less than 1, so the absolute generalization error will tend to increase with $\acc_{val\_same}$. Given that lower precision correlates with lower $\acc_{val\_same}$, Figure \ref{fig:tre_section7} (f), and that lower precision reduces \tre/ reconstruction error, and therefore the values of the \tre/ metric, we hypothesize that low \tre/ values might not be so much an indication of high compositionality so much as an indication of low precision, and thus lower $\acc_{val\_same}$.

We propose a possible approach to correct \tre/ for low expressivity languages could be to simply divide the raw \tre/ score by the precision. We call this metric \textsc{ptre}. Figure \ref{fig:tre_section7} (g) shows a graph of generalization error vs \textsc{ptre}. We can see that \textsc{ptre} actually correlates negatively with generalization error: that is, lower values of \textsc{ptre} correspond to larger generalization error. We hypothesize that this is because higher compositionality (lower \textsc{ptre}) correlates with higher $\acc_{val\_same}$, and thus higher generalization error. Interestingly, a graph comparing generalization error with topologic similarity shows no correlation at all, Figure \ref{fig:tre_section7} (h). We leave further investigation of the relationship between \tre/, \textsc{ptre}, topologic similarity, and underlying ground-truth compositionality to future work.


\section{Related Work}

Our work relates primarily to Shapeworld, \cite{andreas-etal-2018-learning}, which is a dataset of images depicting relations between colored shapes. By comparison with Shapeworld, our work provides a significant larger dataset, and uses textures rather than solid filled shapes, which we would argue allows convolutional networks to learn faster on \sys/, and thus allowed faster experimentation.

Other datasets containing relations include CLEVR \citep{Johnson_2017_CVPR} and CUB-200 \citep{cub200}. CLEVR is a dataset of high-quality high-resolution 3-dimensional images, created using the Blender application \citep{blender} using ray-tracing; along with english language questions, and ground truth answers, which comprise a single english-language word, from a defined vocabulary. By comparison with our work, CLEVR images are much more beautiful, however might require significantly longer training time, since the images are higher resolution, and the objects cannot be recognized by texture alone. Note that CLEVR does incorporate some notion of texture, since objects can be either shiny metal, or matte rubber. CLEVR is a relatively large dataset, comprising 100,000 images, along with around 1 million associated questions. However, \sys/ contains 100k examples, each having 256 images, for a total of about 2.5 million images: significantly larger. In addition, CLEVR targets the visual question-answering setting (`VQA'), and does not clearly map to usage in emergent communications. We experimented with using the CLEVR code to create a dataset for emergent communications, but found that the ray-tracing generation process is relatively slow; and the resulting images are large, and slow to train on.

CUB-200 \citep{cub200} is a dataset of photos of birds, comprising 40-60 images for each of 200 species of birds. English-language annotations were created by \citet{reed2016learning} using Amazon Mechanical Turk, which tasked workers with describing each image in detail, without reference to the class of bird itself. 10 descriptions are available per image. CUB200 along with the annotations from \citet{reed2016learning} was used for example in \citet{mu-etal-2020-shaping}, where it was used as a meta-learning task. However it is not obviously usable for experimentation with meaning space characteristics in an emergent communications setting, since the ground-truth descriptions are free-form text. Free-form text might not be easily usable in metrics such as topographic similarity and \textsc{tre}. In addition the dataset is relatively small, and agents in an emergent communications setting might overfit to training examples. Finally, the images are photos, and might need considerable training time in order for convolutional networks to form effective representations, which might slow down experimentation in an emergent communications setting.

Our work relates also to the earlier SHAPES dataset \citep{andreas2016neural}. SHAPES dataset is similar to the Shapeworld dataset, in that it provides images of solid filled colored shapes. However, like CLEVR, it targets the VQA setting: for each image, english language questions are provided, along with a single binary yes/no response. The SHAPES dataset comprises 64 training images, along with 244 unique questions. By comparison with \sys/, the entire SHAPES training dataset contains four times fewer images than in a single training example from \sys/.

In the domain of emergent communications, there is a plethora of related datasets and tasks, which we can  group primarily into: referential tasks, using images, similar to our own tasks; various complex dynamic scenarios, which are much more complicated than our own tasks; and symbolic tasks, which might be conceptually similar to our own tasks, but use symbolic input.

Looking at referential tasks in existing emergent communications papers, three examples are \citet{lazaridou2018_refgames}, \citet{andreas2017translating}, and \citet{dagan2020co}. \citet{lazaridou2018_refgames} uses a referential task on images of 3d shapes resting on a chequered plane. The images are higher resolution than \sys/, and therefore potentially slower to train. However, each image only contains a single colored object, of differing color and shape. There are no distractor objects, within each image. All objects have the same solid filled texture. There are a total of 4000 images, split across train and test. By comparison with \sys/, the concepts being communicated by the agents are significantly simpler, i.e. a single color and shape, and the dataset is much smaller. The absence of distractor objects within each image potentially might lead to the agent using simpler strategies than we would like, e.g. only communicating the color, and not the shape. \citet{andreas2017translating} uses the CUB-200 dataset we mentioned earlier. \citet{dagan2020co} uses a modified version of the SHAPES dataset. \cite{dagan2020co} creates a dataset having 80k training examples, which is similar to \sys/ in number of training examples. However, each example comprises a single colored shape, without distractor objects, so the task is potentially much simpler than our own tasks.

As far as more complex dynamic emergent communications scenarios, some examples are the Driving task from \citet{andreas2017translating}, the Traffic Junction task and the Combat task from \citet{sukhbaatar_fergus_multiagent_communication}, and \citet{mordatch2018emergence}. These tasks take place in dynamic 2d game worlds, which might be challenging for the agent to learn. Whilst these tasks are exciting and interesting, given the complex dynamic nature of these tasks, from the point of view of investigation into the generated language, the tasks might be difficult to reason over. Credit assigment to different aspects of the task for aspects of the resulting language might be challenging. For example, it might be unclear what attributes are being communicated by an agent at each time step. We believe that it might be interesting in a first time to experiment with more restricted supervised datasets, where we have close control over the attributes that the agents need to communicate.

Examples of emergent communications works using symbolic input include \citet{kottur2017natural}, \citet{lowe2019pitfalls}, \citet{ren2020compositional}, \citet{li2019ease}, \citet{slowik2020exploring}, and \citet{andreas_measuring_compositionality_2019}. \citet{kottur2017natural} uses symbolic representations of shapes, colors and styles, in a multi-turn referential task setting. \citet{lowe2019pitfalls} uses matrix communication games to investigate aspects of emergent communications. Matrix communication games are a generalization of games such as the Prisoner's Dilemna \citep{sally1995conversation} to randomly generated pay-off matrices. \citet{ren2020compositional}, \citet{li2019ease} and \citet{andreas_measuring_compositionality_2019} use a referential task with symbolic inputs, similar to \citet{kottur2017natural}. Lastly, \citet{slowik2020exploring} uses a referential task for symbolic inputs comprising hierarchical concepts. We hypothesize that these works used symbolic inputs because it is the most computationally efficient approach. However, for the investigation of the form of emergent languages, including analysis of compositionality, we believe that it might be interesting to run experiments such as these using non-symbolic input, so that it is clear the compositional output is not simply a reflection of the compositional symbolic input. In our own work, we carried out a case study of using \sys/ to reproduce section 7 of \citet{andreas_measuring_compositionality_2019}. The entire section 7 experimentation ran on a single consumer NVIDIA 2080Ti GPU in just over a day, which we feel might be a relatively modest training budget. We hope that the creation of \sys/ might provide one additional possible option for future works which could take advantage of computationally efficient training, on non-symbolic inputs.



\section{Conclusion}

We have presented \sys/ a dataset targeted at fast, green experimentation for emergent communication. \sys/ provides a number of challenging tasks, such as relations learning, and with tight negatives, to minimize any tendancy for agents to find `short-cuts'. The dataset images are relatively low-resolution, by using textures rather than solid outline shapes, to differentiate between different objects. We provided extensive experimentation on \sys/. We compared \sys/ with an alternative relations dataset, `Shapeworld', and showed that in an emergent communications scenario \sys/ led to rapid learning, whilst being a challenging task, performing nevertheless far from 100\% accuracy. We experimented with different sender and receiver model architectures, finding that StackedInputs and Prototypical sender worked well; and for the receiver, and Cosine receiver worked well as a receiver model. For multi-task learning, we found that adding in additional tasks whilst training on specific target task did not tend to improve the accuracy, however it did show marked improvement in the measured topological rho compositionality metric. We carried out an experiment on the extent to which increasing the size of the meaning space led to higher compositionality. Surprisingly we found that this was not the case. We hypothesized that topological similarity decreasing with the size of the meaning space might reflect a limitation of the use of topological similarity to measure underlying compositionality more than that the increasing meaning space actually decreases underlying ground-truth compositionality. Lastly, we carried our a case-study of replacing a symbolic dataset in an existing work with the \sys/ dataset. We found that the entire experiment could could be run on a single consumer GPU, using \sys/, in less than a day.

We hope that \sys/ can provide a helpful experimental playground that is an alternative to using symbolic data, and that allows for fast, green experimentation on emergent communications.

\section*{Acknowledgements}

Thank you to Angeliki Lazaridou for many interesting discussions and ideas.



\bibliography{ms}

\begin{thebibliography}{29}
\expandafter\ifx\csname natexlab\endcsname\relax\def\natexlab#1{#1}\fi

\bibitem[{Andreas(2019)}]{andreas_measuring_compositionality_2019}
Jacob Andreas. 2019.
\newblock \href {https://openreview.net/forum?id=HJz05o0qK7} {Measuring
  compositionality in representation learning}.
\newblock In \emph{International Conference on Learning Representations}.

\bibitem[{Andreas et~al.(2017)Andreas, Dragan, and
  Klein}]{andreas2017translating}
Jacob Andreas, Anca Dragan, and Dan Klein. 2017.
\newblock \href {https://doi.org/10.18653/v1/P17-1022} {Translating neuralese}.
\newblock In \emph{Proceedings of the 55th Annual Meeting of the Association
  for Computational Linguistics (Volume 1: Long Papers)}, pages 232--242,
  Vancouver, Canada. Association for Computational Linguistics.

\bibitem[{Andreas et~al.(2018)Andreas, Klein, and
  Levine}]{andreas-etal-2018-learning}
Jacob Andreas, Dan Klein, and Sergey Levine. 2018.
\newblock \href {https://doi.org/10.18653/v1/N18-1197} {Learning with latent
  language}.
\newblock In \emph{Proceedings of the 2018 Conference of the North {A}merican
  Chapter of the Association for Computational Linguistics: Human Language
  Technologies, Volume 1 (Long Papers)}, pages 2166--2179, New Orleans,
  Louisiana. Association for Computational Linguistics.

\bibitem[{Andreas et~al.(2016)Andreas, Rohrbach, Darrell, and
  Klein}]{andreas2016neural}
Jacob Andreas, Marcus Rohrbach, Trevor Darrell, and Dan Klein. 2016.
\newblock Neural module networks.
\newblock In \emph{Proceedings of the IEEE conference on computer vision and
  pattern recognition}, pages 39--48.

\bibitem[{{Blender Foundation}(2002)}]{blender}
{Blender Foundation}. 2002.
\newblock {Blender}.
\newblock \url{https://www.blender.org/}.
\newblock Accessed: 2021-05-22.

\bibitem[{Brighton and Kirby(2006)}]{brighton_kirby_2006_topographic_mappings}
Henry Brighton and Simon Kirby. 2006.
\newblock Understanding linguistic evolution by visualizing the emergence of
  topographic mappings.
\newblock \emph{Artificial life}, 12(2):229--242.

\bibitem[{Cao et~al.(2018)Cao, Lazaridou, Lanctot, Leibo, Tuyls, and
  Clark}]{cao2018emergent}
Kris Cao, Angeliki Lazaridou, Marc Lanctot, Joel~Z Leibo, Karl Tuyls, and
  Stephen Clark. 2018.
\newblock Emergent communication through negotiation.
\newblock \emph{arXiv preprint arXiv:1804.03980}.

\bibitem[{Chaplot et~al.(2018)Chaplot, Mysore~Sathyendra, Pasumarthi,
  Rajagopal, and Salakhutdinov}]{Chaplotetal2018}
Devendra~Singh Chaplot, Kanthashree Mysore~Sathyendra, Rama~Kumar Pasumarthi,
  Dheeraj Rajagopal, and Ruslan Salakhutdinov. 2018.
\newblock \href {https://ojs.aaai.org/index.php/AAAI/article/view/11832}
  {Gated-attention architectures for task-oriented language grounding}.
\newblock \emph{Proceedings of the AAAI Conference on Artificial Intelligence},
  32(1).

\bibitem[{Chen et~al.(2015)Chen, Wang, Chen, Gao, Xu, and
  Nevatia}]{chen2015abc}
Kan Chen, Jiang Wang, Liang-Chieh Chen, Haoyuan Gao, Wei Xu, and Ram Nevatia.
  2015.
\newblock Abc-cnn: An attention based convolutional neural network for visual
  question answering.
\newblock \emph{arXiv preprint arXiv:1511.05960}.

\bibitem[{Dagan et~al.(2020)Dagan, Hupkes, and Bruni}]{dagan2020co}
Gautier Dagan, Dieuwke Hupkes, and Elia Bruni. 2020.
\newblock Co-evolution of language and agents in referential games.
\newblock \emph{arXiv preprint arXiv:2001.03361}.

\bibitem[{Johnson et~al.(2017)Johnson, Hariharan, van~der Maaten, Fei-Fei,
  Lawrence~Zitnick, and Girshick}]{Johnson_2017_CVPR}
Justin Johnson, Bharath Hariharan, Laurens van~der Maaten, Li~Fei-Fei,
  C.~Lawrence~Zitnick, and Ross Girshick. 2017.
\newblock Clevr: A diagnostic dataset for compositional language and elementary
  visual reasoning.
\newblock In \emph{Proceedings of the IEEE Conference on Computer Vision and
  Pattern Recognition (CVPR)}.

\bibitem[{Khurshudov(2015)}]{leopardspotsofa}
Artem Khurshudov. 2015.
\newblock Suddenly a leopard print sofa appears.
\newblock \url{https://archive.is/PhExN}.
\newblock Accessed: 2021-05-22.

\bibitem[{Kirby et~al.(2015)Kirby, Tamariz, Cornish, and
  Smith}]{kirby2015compression}
Simon Kirby, Monica Tamariz, Hannah Cornish, and Kenny Smith. 2015.
\newblock Compression and communication in the cultural evolution of linguistic
  structure.
\newblock \emph{Cognition}, 141:87--102.

\bibitem[{Kottur et~al.(2017)Kottur, Moura, Lee, and Batra}]{kottur2017natural}
Satwik Kottur, Jos{\'e} Moura, Stefan Lee, and Dhruv Batra. 2017.
\newblock \href {https://doi.org/10.18653/v1/D17-1321} {Natural language does
  not emerge {`}naturally{'} in multi-agent dialog}.
\newblock In \emph{Proceedings of the 2017 Conference on Empirical Methods in
  Natural Language Processing}, pages 2962--2967, Copenhagen, Denmark.
  Association for Computational Linguistics.

\bibitem[{Lazaridou et~al.(2018)Lazaridou, Hermann, Tuyls, and
  Clark}]{lazaridou2018_refgames}
Angeliki Lazaridou, Karl~Moritz Hermann, Karl Tuyls, and Stephen Clark. 2018.
\newblock \href {https://openreview.net/forum?id=HJGv1Z-AW} {Emergence of
  linguistic communication from referential games with symbolic and pixel
  input}.
\newblock In \emph{International Conference on Learning Representations}.

\bibitem[{Li and Bowling(2019)}]{li2019ease}
Fushan Li and Michael Bowling. 2019.
\newblock Ease-of-teaching and language structure from emergent communication.
\newblock \emph{arXiv preprint arXiv:1906.02403}.

\bibitem[{Locatello et~al.(2019)Locatello, Bauer, Lucic, Raetsch, Gelly,
  Sch{\"o}lkopf, and Bachem}]{locatello2019challenging}
Francesco Locatello, Stefan Bauer, Mario Lucic, Gunnar Raetsch, Sylvain Gelly,
  Bernhard Sch{\"o}lkopf, and Olivier Bachem. 2019.
\newblock Challenging common assumptions in the unsupervised learning of
  disentangled representations.
\newblock In \emph{international conference on machine learning}, pages
  4114--4124. PMLR.

\bibitem[{Lowe et~al.(2019)Lowe, Foerster, Boureau, Pineau, and
  Dauphin}]{lowe2019pitfalls}
Ryan Lowe, Jakob Foerster, Y-Lan Boureau, Joelle Pineau, and Yann Dauphin.
  2019.
\newblock On the pitfalls of measuring emergent communication.
\newblock In \emph{International Conference on Autonomous Agents and Multiagent
  Systems (AAMAS)}.

\bibitem[{Misra et~al.(2017)Misra, Langford, and Artzi}]{misra2017mapping}
Dipendra Misra, John Langford, and Yoav Artzi. 2017.
\newblock Mapping instructions and visual observations to actions with
  reinforcement learning.
\newblock \emph{arXiv preprint arXiv:1704.08795}.

\bibitem[{Mordatch and Abbeel(2018)}]{mordatch2018emergence}
Igor Mordatch and Pieter Abbeel. 2018.
\newblock Emergence of grounded compositional language in multi-agent
  populations.
\newblock In \emph{Proceedings of the AAAI Conference on Artificial
  Intelligence}, volume~32.

\bibitem[{Mu et~al.(2020)Mu, Liang, and Goodman}]{mu-etal-2020-shaping}
Jesse Mu, Percy Liang, and Noah Goodman. 2020.
\newblock \href {https://doi.org/10.18653/v1/2020.acl-main.436} {Shaping visual
  representations with language for few-shot classification}.
\newblock In \emph{Proceedings of the 58th Annual Meeting of the Association
  for Computational Linguistics}, pages 4823--4830, Online. Association for
  Computational Linguistics.

\bibitem[{Reed et~al.(2016)Reed, Akata, Lee, and Schiele}]{reed2016learning}
Scott Reed, Zeynep Akata, Honglak Lee, and Bernt Schiele. 2016.
\newblock Learning deep representations of fine-grained visual descriptions.
\newblock In \emph{Proceedings of the IEEE conference on computer vision and
  pattern recognition}, pages 49--58.

\bibitem[{Ren et~al.(2020)Ren, Guo, Labeau, Cohen, and
  Kirby}]{ren2020compositional}
Yi~Ren, Shangmin Guo, Matthieu Labeau, Shay~B Cohen, and Simon Kirby. 2020.
\newblock Compositional languages emerge in a neural iterated learning model.
\newblock \emph{arXiv preprint arXiv:2002.01365}.

\bibitem[{Sally(1995)}]{sally1995conversation}
David Sally. 1995.
\newblock Conversation and cooperation in social dilemmas: A meta-analysis of
  experiments from 1958 to 1992.
\newblock \emph{Rationality and society}, 7(1):58--92.

\bibitem[{Shi et~al.(2015)Shi, Chen, Wang, Yeung, kin Wong, and chun
  Woo}]{shi2015convolutional}
Xingjian Shi, Zhourong Chen, Hao Wang, Dit-Yan Yeung, Wai kin Wong, and Wang
  chun Woo. 2015.
\newblock \href {http://arxiv.org/abs/1506.04214} {Convolutional lstm network:
  A machine learning approach for precipitation nowcasting}.

\bibitem[{S{\l}owik et~al.(2020)S{\l}owik, Gupta, Hamilton, Jamnik, Holden, and
  Pal}]{slowik2020exploring}
Agnieszka S{\l}owik, Abhinav Gupta, William~L Hamilton, Mateja Jamnik, Sean~B
  Holden, and Christopher Pal. 2020.
\newblock Exploring structural inductive biases in emergent communication.
\newblock \emph{arXiv preprint arXiv:2002.01335}.

\bibitem[{Snell et~al.(2017)Snell, Swersky, and Zemel}]{snell2017prototypical}
Jake Snell, Kevin Swersky, and Richard~S Zemel. 2017.
\newblock Prototypical networks for few-shot learning.
\newblock \emph{arXiv preprint arXiv:1703.05175}.

\bibitem[{Sukhbaatar et~al.(2016)Sukhbaatar, Szlam, and
  Fergus}]{sukhbaatar_fergus_multiagent_communication}
Sainbayar Sukhbaatar, Arthur Szlam, and Rob Fergus. 2016.
\newblock \href {http://arxiv.org/abs/1605.07736} {Learning multiagent
  communication with backpropagation}.
\newblock \emph{CoRR}, abs/1605.07736.

\bibitem[{Welinder et~al.(2010)Welinder, Branson, Mita, Wah, Schroff, Belongie,
  and Perona}]{cub200}
Peter Welinder, Steve Branson, Takeshi Mita, Catherine Wah, Florian Schroff,
  Serge Belongie, and Pietro Perona. 2010.
\newblock Caltech-ucsd birds 200.

\end{thebibliography}
\bibliographystyle{acl_natbib}


\end{document}